\pdfoutput=1
\documentclass[11pt]{article}
\usepackage{acl}

\usepackage{times}
\usepackage{latexsym}
\usepackage[T1]{fontenc}
\usepackage[utf8]{inputenc}
\usepackage{microtype}
\usepackage{tabularx}
\usepackage{microtype}
\usepackage{multirow}
\usepackage{array}
\usepackage{booktabs}
\usepackage{paralist}
\usepackage{graphicx}
\usepackage{xcolor}
\usepackage{xspace}
\usepackage{hyperref}
\usepackage{url}
\usepackage{xspace}
\usepackage{amsmath}
\usepackage{float}
\DeclareMathOperator*{\argmax}{arg\,max}

\newcommand{\F}{$\textrm{F}_1$\xspace}
\newcommand{\emonamme}{Emo-Name\xspace}
\newcommand{\expremo}{Expr-Emo\xspace}
\newcommand{\feelsemo}{Feels-Emo\xspace}
\newcommand{\wordnet}{WN-Def\xspace}
\newcommand{\emosyn}{Emo-S\xspace}
\newcommand{\exprsyn}{Expr-S\xspace}
\newcommand{\feelsyn}{Feels-S\xspace}
\newcommand{\emolex}{EmoLex\xspace}
\newcommand{\isear}{\textsc{Isear}\xspace}
\newcommand{\blogs}{\textsc{Blogs}\xspace}
\newcommand{\tec}{\textsc{Tec}\xspace}

\title{Natural Language Inference Prompts for\\ Zero-shot Emotion Classification in Text across Corpora}

\author{\bf {Flor Miriam} {Plaza-del-Arco}$^{1,2}$, \bf {María-Teresa} {Martín-Valdivia}$^1$, \bf Roman Klinger$^2$\\
  $^1$SINAI, Computer Science Department, CEATIC, Universidad de Ja{\'e}n, Spain \\
  $^2$Institut für Maschinelle Sprachverarbeitung, University of Stuttgart, Germany\\ \texttt{\{fmplaza, maite\}@ujaen.es},\hspace{1em} \texttt{klinger@ims.uni-stuttgart.de}}

\begin{document}
\maketitle
\begin{abstract}
  Within textual emotion classification, the set of relevant labels
  depends on the domain and application scenario and might not be
  known at the time of model development. This conflicts with the
  classical paradigm of supervised learning in which the labels need
  to be predefined. A solution to obtain a model with a flexible set
  of labels is to use the paradigm of zero-shot learning as a natural
  language inference task, which in addition adds the advantage of not
  needing any labeled training data.
  This raises the question how to prompt a natural language inference
  model for zero-shot learning emotion classification. Options for
  prompt formulations include the emotion name \textit{anger} alone or
  the statement ``This text expresses \textit{anger}''. With this
  paper, we analyze how sensitive a natural language inference-based
  zero-shot-learning classifier is to such changes to the prompt under
  consideration of the corpus: How carefully does the prompt need to
  be selected?
  We perform experiments on an established set of emotion datasets
  presenting different language registers according to different
  sources (tweets, events, blogs) with three natural language
  inference models and show that indeed the choice of a particular
  prompt formulation needs to fit to the corpus. We show that
  this challenge can be tackled with combinations of multiple
  prompts. Such ensemble is more robust across corpora than individual
  prompts and shows nearly the same performance as the individual best
  prompt for a particular corpus.
\end{abstract}

\begin{figure}[b!]
\centering
\includegraphics[width=0.47\textwidth]{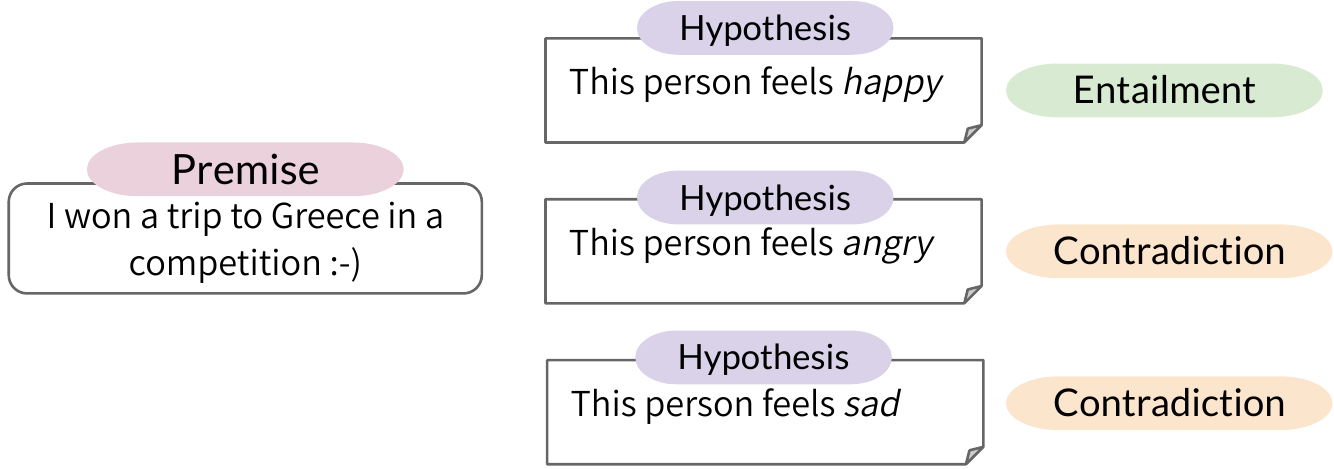}
\caption{An example of the application of NLI to ZSL emotion
  classification. Given the premise ``I won a trip to Greece in a
  competition'', three hypotheses represent the emotions
  (\textit{joy}, \textit{anger}, \textit{sadness}). The representation
  of \textit{joy} is entailed and therefore predicted.}
\label{fig:NLI_example}
\end{figure}

\section{Introduction}
To enable communication about emotions, there exists a set of various
emotion names, for instance those labeled as \emph{basic emotions}, by
\citet{ekman1992argument} or \citet{plutchik2001nature}
(\textit{anger}, \textit{fear}, \textit{joy}, \textit{sadness},
\textit{disgust}, \textit{surprise}, \textit{trust},
\textit{anticipation}). While such psychological models influence
natural language processing and emotion categorization approaches, the
choice of emotion concepts is context-dependent. For instance,
\citet{Scherer1997} and \citet{Troiano2019} opted to use
\textit{guilt} and \textit{shame} as self-directed emotions in
addition to Ekman's basic emotions, to analyze self-reports of
events. For the context of the perception of art it is more
appropriate to consider \textit{aesthetic emotions}
\citep{Menninghaus2019, haider-etal-2020-po}, like \textit{beauty},
\textit{sublime}, \textit{inspiration}, \textit{nostalgia}, and
\textit{melancholia}.

This leads to a potential gap between concepts in emotion-related
training data and the application domain, purely because the label set
is not compatible. One solution is to resort to so-called dimensional
models, in which emotion names are located in vector spaces of affect
\citep[valence,
arousal,][]{preotiuc-pietro-etal-2016-modelling, buechel-hahn-2017-emobank}
or cognitive appraisal \citep[e.g., regarding \textit{responsibility},
\textit{certainty}, \textit{pleasantness}, \textit{control},
\textit{attention} with respect to a stimulus event,
][]{Hofmann2020,Troiano2023}. In these vector spaces, classes can be
assigned to predicted points with a nearest-neighbor approach, even if
these classes have not been seen during training. This approach,
however, has the disadvantage of the so-called hubness problem
\citep{Lazaridou2015}, namely that the distance between predictions
and concepts that have been seen during training tends to be smaller
than to novel concepts.  We acknowledge ongoing research to tackle
this problem \citep{park-etal-2021-dimensional,Buechel2021}.

We take a different, more direct, route to obtaining classifiers for
discrete emotion categories which are not known at system development
time, namely zero-shot learning (ZSL). One instantiation of such ZSL
systems is via natural language inference models (ZSL-NLI), in which
the inference task needs to perform reasoning \citep{Yin2019}.
Consequently, the idea of implementing ZSL-NLI models is not by
exemplification and optimizing a classifier, but developing
appropriate natural language class name representations which we refer
to as \textit{prompts}. We see an example for the application of an
NLI model to ZSL emotion classification in
Figure~\ref{fig:NLI_example} -- the NLI model needs to decide if the
hypothesis (a prompt which represents the class label) entails the
premise (which corresponds to the instance to be classified).  This
paradigm raises the question (which we answer in this paper) of how to
formulate the emotion prompt and how much the design choice of the
prompt needs to fit the dataset.

Manually developing intuitive templates based on human data
introspection may be the most natural method to produce prompts. In
this paper, we provide manually created templates to probe emotion
classification in an NLI-ZSL setup and we analyze whether prompts are
language-register dependent according to various corpora (tweets,
event descriptions, blog posts). To accomplish this aim, we perform
experiments on an established set of emotion datasets with three NLI
models and we show that (1) prompts are indeed corpus-specific and
that the differences follow the same pattern across different
pretrained NLI models, (2) that an ensemble of multiple prompts
behaves more robustly across corpora, and (3) the representation of
the emotion concept as part of the textual prompt is an important
element, benefiting from representations with synonyms and related
concepts, instead of just the emotion name. Our code is publicly
available at \url{https://github.com/fmplaza/zsl_nli_emotion_prompts}.

\section{Related Work}

\subsection{Emotion Classification}

Emotion analysis has become a major area of research in NLP which
comprises a variety of tasks, including emotion stimulus or cause
detection \citep{li-etal-2021-boundary, DoanDang2021} and emotion
intensity prediction \citep{mohammad-bravo-marquez-2017-emotion,
  koper-etal-2017-ims}. The task of emotion classification received
most attention in recent years
\citep[i.a.]{bostan-klinger-2018-analysis, mohammad-etal-2018-semeval,
  plaza-del-arco-etal-2020-emoevent}.

Emotion classification aims at mapping textual units to an emotion
category. The categories often rely on psychological theories such as
those proposed by \citet{ekman1992argument} (\textit{anger},
\textit{fear}, \textit{sadness}, \textit{joy}, \textit{disgust},
\textit{surprise}), or the dimensional model of
\citet{plutchik2001nature} (adding \textit{trust} and
\textit{anticipation}). However, neither are all these basic
emotions relevant in all domains, nor are they sufficient.  For
instance, in the education field, \citet{d2007mind} found
\textit{boredom}, \textit{confusion}, \textit{flow},
\textit{frustration}, and \textit{delight} to be more relevant than
\textit{fear} or \textit{disgust}. \citet{sreeja2015applying} reveal
that emotions such as \textit{love}, \textit{hate}, and
\textit{courage} are necessary to model the emotional perception of
poetry. \citet{bostan-etal-2020-goodnewseveryone} identify
\textit{annoyance}, \textit{guilt}, \textit{pessimism}, or
\textit{optimism} to be important to analyze news headlines.

A strategy to avoid specification of discrete categories is the use of
dimensional spaces that consider valence, arousal, and dominance
\citep[VAD, ][]{russell1977evidence}. \citet{smith1985patterns} claim
that this model does not represent important difference between
emotions and propose an alternative dimensional model based on
cognitive appraisal, which has recently been used for text analysis
\citep{Hofmann2020,stranisci-EtAl:2022:LREC,Troiano2023}.
Independent of the classification or regression approach, nearly all
recently proposed systems rely on transfer learning from general
language representations. We refer the reader to recent shared task
surveys for a more comprehensive overview
\citep{Mohammad2018,tafreshi-etal-2021-wassa, plaza2021emoevales}.

\subsection{Zero-shot Learning}

Zero-shot learning (ZSL) aims at performing predictions without having
seen labeled training examples specific for the concrete
task. Zero-shot methods typically work by associating seen and unseen
classes using auxiliary information, which encodes observable
distinguishing properties of instances \citep{8413121}. In NLP, the
term is used predominantly either to refer to cross-lingual transfer to
languages that have not been seen at training time (change of the
language), or to predict classes that have not been seen \citep[change of the
labels, ][]{10.1145/3293318}. Our work falls in the second category.

Various approaches exist to perform zero-shot text
classification. One approach represents labels in an embedding space
\citep[i.a.]{2999611.2999716, esann16zeroshot,
  rios-kavuluru-2018-shot}. A model is trained to predict the
respective embedding vectors for categorical labels. At test time,
embeddings of novel labels need to be known and will be assigned if
the distance between the predicted embedding and the label embedding
is small.  This method suffers from the hubness problem, that is, when
the semantic label embeddings are close to each other, the projection
of labels to the semantic space forms hubs \citep{radovanovic10a}.

Another approach is to use transformer language models to classify if
a label embedding is compatible with an instance embedding
\citep{BrownMRSKDNSSAA20}. To this end, no labeled examples are
provided at training phase but an instruction in natural language is
given to the model to interpret the label class (the
\textit{prompt}). An instance of this approach is \emph{Task-Aware
  Representations} \citep[TARS, ][]{halder-etal-2020-task} who
separate the instance text and the class label by the special
separator token \texttt{[SEP]} in BERT \citep{Devlin2019}.

An alternative is to treat ZSL as textual entailment. Following this
approach, \citet{Yin2019} propose a sequence-pair classifier that
takes two sentences as input (a premise and a hypothesis) and decides
whether they entail or contradict each other. They study various
formulations of the labels as hypotheses and evaluate the method in
various NLP tasks including topic detection, situation detection, and
emotion classification.  In their evaluation, emotion classification
turns out to be most challenging. Another study that conducted prompt
engineering in NLI models proposes probabilistic ZSL ensembles for
emotion classification \citep{basile-etal-2021-probabilistic}. The
authors experiment with the same prompts as \citet{Yin2019} and
aggregate the predictions of multiple NLI models using Multi-Annotator
Competence Estimation (MACE), a method developed for modelling
crowdsourced annotations.

Our work on ZSL for emotion classification differs from previous
approaches as follows. We analyze whether prompts are corpus-specific
and propose an ensemble of multiple prompts to achieve a classifier
which is more robust across corpora (in contrast to an
ensemble of multiple NLI models in the work by
\citet{basile-etal-2021-probabilistic}). Further, we analyze if the
introduction of more knowledge about the emotion in the prompt through
emotion synonyms and related concepts helps its interpretation in the
NLI models.

\begin{table*}[t]
\centering
\small
\begin{tabular}{lll}
\toprule
ID   &   Prompt   & Example \\
\cmidrule(l){1-1} \cmidrule(l){2-2} \cmidrule(l){3-3}
\emonamme & \textit{emotion name} & \textit{joy} \\
\expremo & This text expresses \textit{emotion name} & This text expresses \textit{joy} \\
\feelsemo & This person feels \textit{emotion name} & This person feels \textit{joyful} \\
\wordnet & This person expresses \textit{WordNet def.} &  This person expresses \textit{a feeling of great pleasure and happiness}\\
\midrule
\emosyn & \textit{emotion synonym} & \textit{happy} \\
\exprsyn & This text expresses  e\textit{motion syn.} & This text expresses \textit{happiness} \\
\feelsyn & This person feels  \textit{emotion syn.}  & This person feels  \textit{happy} \\
\midrule
\emolex & \textit{emotion lexicon} & \textit{party} \\
\bottomrule
\end{tabular}
    \caption{Emotion prompts. Words in \textit{italics} represent
      placeholders for the emotion concept representation.}
    \label{tab:prompts}
\end{table*}

\section{Methods}\label{sec:methods}
In this section, we explain how we apply NLI for ZSL emotion
classification and propose a collection of prompts to contextualize
and represent the emotion concept in different corpora. In addition,
we propose a prompt ensemble which is more robust across corpora.

\subsection{Natural Language Inference for Zero-shot Emotion
  Classification}
\label{sec:nli_emotion}

The NLI task is commonly defined as a sentence-pair classification in
which two sentences are given: a \textit{premise} $s_1$ and a
\textit{hypothesis} $s_2$. The task is to learn a
function $f_{\mathrm{NLI}}(s_1,s_2) \rightarrow \{E,C,N\}$, in which
$E$ expresses the entailment of $s_1$ and $s_2$, $C$ denotes a
contradiction and $N$ is a neutral output.

We treat ZSL emotion classification as a textual entailment problem,
but represent each label under consideration with multiple prompts, in
contrast to \citet{Yin2019}. Given a sentence to be classified $x$
(\textit{premise}) and an emotion $e$, we have a function $g(e)$ that
generates a set of prompts (\textit{hypothesis}) out of the class
$e\in E$ (with $E$ being the set of emotions under
consideration). Under the assumption of an NLI model $m$, which
calculates the entailment probability $p_m(\gamma, x)$ for some
emotion representation $\gamma \in g(e)$, we assign the average
entailment probability across all emotion representations as
\[\bar{p}^g_m(e,x)=\frac{1}{|g(e)|} \sum_{\gamma\in g(e)} p_m(\gamma, x)\] for
a particular prompt generation method $g$. The classification
decision \[\hat{e}^g_x = \argmax_{e\in E} \bar{p}^g_m(e,x)\] returns
the emotion corresponding to the maximum entailment probability.

\subsection{Emotion Prompts}

In the context of emotion analysis, two important questions arise when
formulating a prompt: (\textit{i})~How to contextualize the emotion
name, and (\textit{ii}) How to represent the emotion concept?

\subsubsection{Prompt Generation}\label{sec:prompt_gen}

We generate a set of prompts with the function $g(e) = c + r(e)$, in
which $c$ represents what we call the \emph{context} and $r(e)$
represents a set of emotion representations.\footnote{In principle,
  $c$ could also be a set. $g(e)$ would then need to use a
  cross-product instead of the element-wise concatenation $+$, which
  we use in our experiments.}  As $c$, we use either an empty string
$\epsilon$, the text ``\textit{This text expresses}'', ``\textit{This
  person feels}'', or ``\textit{This person expresses}'', motivated by
our choice of the language register presented in the datasets used in our experiments (see
\S~\ref{sec:experiments}).

\subsubsection{Prompts for Zero-Shot Emotion Classification}
\label{sec:emo_prompts}
Each prompt in this paper consists of context and the emotion
representation. There are three prompts which have in common the
emotion name representation, namely \emonamme, \expremo, and
\feelsemo. Variations of these prompts are \emosyn, \exprsyn, and
\feelsyn, where the emotion name representation is replaced by
multiple emotion synonyms and \emolex where the emotion name is
replaced by entries from an emotion word lexicon. In detail, we use
the following prompts (Table~\ref{tab:prompts} shows examples):
\begin{compactdesc}
\item[\textbf{\emonamme}.] $c=\epsilon$ and $r(e) = \{e\}$.
\item[\textbf{\expremo}.] $c$ = $\textrm{``This text expresses''}$, $r(e)$ = $\{e\}$.
\item[\textbf{\feelsemo}.] $c$ = $\textrm{``This person feels''}$, $r(e)$ = $\{e\}$.
\item[\textbf{\wordnet}.] $c$ = $\textrm{``This person expresses''}$ and
  $r(e) = \{\textrm{WN-Def}(e)\}$, where $\textrm{WN-Def}(e)$ is the
  WordNet definition for $e$ \citep{10.1145/219717.219748}.
\item[\textbf{\emosyn}.] We aim to see whether incorporating
  additional information using a set of abstract emotion-related names
  leads to a better model. Hence, we set $r(e)$ to return a set of
  emotion synonyms for $e$. Table \ref{tab:emo_syn} shows the emotion
  synonyms considered for each emotion.\footnote{Each synonym is
    grammatically adapted for the context of the prompts \exprsyn and
    \feelsyn.}
\item[\textbf{\exprsyn}.] We set $r(e)$ to be the same as in \emosyn,
  but additionally set $c=\textrm{``This text
    expresses''}$. Therefore, $g(e)$ returns all combinations of this
  string with each synonym.
\item[\textbf{\feelsyn}.] This prompt is the same as \exprsyn with the
  difference that we set $c=\textrm{``This person feels''}$.
\item[\textbf{\emolex}.] This prompt is different from the previous
  ones, which consisted of small sets of context/emotion
  representation combinations. Here, $c = \epsilon$ , but for the
  emotion representation we use a large popular lexicon, namely Emolex
  \cite{mohammad2013crowdsourcing} to assign all entries associated
  with $e$ in this lexicon. This generates prompts which contain
  abstract emotion synonyms as well as concrete objects (like
  \textit{gift} for \textit{joy}).
\end{compactdesc}

\subsection{Ensemble of prompts}
\label{sec:ensemble_prompts}
In practical applications, the choice of a particular prompt could not
be performed manually by some user. Under the assumption that the
choice of prompts is indeed corpus-specific, we combine multiple
prompt sets in an ensemble.

The ensemble model takes as input a text $x$ and a set of 
prompt-generating models $G$ with $\bar{p}^g_M(e,x)$ ($g\in G$).
The ensemble decision is then
\[
\hat{e}(x,m) = \argmax_{e\in E}\frac{1}{|G|}\sum_{g\in G}\bar{p}^g_m(e,x)\,.
\]

\section{Experiments}
\label{sec:experiments}

We aim at answering the following research questions: \textbf{(RQ1)}
Do NLI models behave the same across prompts?  \textbf{(RQ2)} Should
we use synonyms for the emotion representation?  \textbf{(RQ3)} Is an
ensemble of multiple prompts more robust across corpora?
\textbf{(RQ4)} Are synonyms sufficient? Would it be even more useful
to use more diverse representations of emotions?

\subsection{Experimental Setting}

\subsubsection{Datasets}
\label{sec:datasets}
We compare our methods on three English corpora, to gain an
understanding of the role of the respective corpus. \textsc{Tec}
\citep{mohammad2012emotional} contains 21,051 tweets weakly labeled
according to hashtags corresponding to the six Ekman emotions
\citep{ekman1992argument}: \textit{\#anger}, \textit{\#disgust},
\textit{\#fear}, \textit{\#happy}, \textit{\#sadness}, and
\textit{\#surprise}. \textsc{Isear} \citep{Scherer1997} includes 7,665
English self-reports of events that triggered one of the emotions
(\textit{joy}, \textit{fear}, \textit{anger}, \textit{sadness},
\textit{disgust}, \textit{shame}, and \textit{guilt}). \textsc{Blogs}
\citep{aman2007identifying} consists of 5,205 sentences from 173 blogs
compiled from the Web using a list of emotion-related seed words. It
is human-annotated according to Ekman's set of basic emotions and an
additional \textit{no emotion} category. \tec and \isear are publicly
available for research purposes and \blogs is available upon
request. All datasets are anonymized by the authors.

\begin{table}[t]
\centering
\small
\setlength{\tabcolsep}{3pt} 
\begin{tabular}{llrll}
  \toprule
  Dataset          &  Labels      & Size      & Source & Avail. \\
  \cmidrule(l){1-1} \cmidrule(l){2-2} \cmidrule(l){3-3} \cmidrule(l){4-4}
  \cmidrule(l){5-5}
  \textsc{Tec}                   & Ekman                   &  21,051   &  tweets & D-RO     \\
  \textsc{Blogs}                 & Ekman $+$ \textit{no emotion}      &  5,205    & blogs & R        \\
  \textsc{Isear}                 & Ekman $-$ \textit{Su} + \textit{G} +
                                   \textit{Sh}
                                  & 15,302    &   events & GPLv3 \\
  \bottomrule
\end{tabular}
\caption{Datasets used in our experiments (Su: surprise, G: guilt, Sh: shame) [D-RO] available to download, research only, [R] Available upon request, [GPLv3] GNU Public License version 3.}
\label{tab:datasets}
\end{table}

These corpora differ in various parameters (see Table
\ref{tab:datasets}): the annotation scheme (variations of Ekman's
model), the corpus source (tweets, events, blogs), the annotation
procedure (hashtag, crowdsourcing, self-reporting), and the size. Note
that the annotation procedure that the ZSL method needs to reconstruct
varies in complexity.

\subsubsection{NLI Models and Baseline}
\label{sec:models}
We compare our ZSL models with an empirical upper bound, namely a
RoBERTa model fine-tuned with supervised training
\citep{liu2019roberta} on each emotion dataset described in
\S~\ref{sec:emo_prompts}. We fine-tune RoBERTa for three epochs, the
batch size is set to 32 and the learning rate to $2\cdot10^{-5}$. No
hyperparameter search has been applied. We perform 10-fold
cross-validation and report the results on the whole data set (as we
do with the NLI models).

For our ZSL experiments, we explore three state-of-the-art pretrained
NLI models publicly available within the Hugging Face Transformers
Python library \citep{wolf-etal-2020-transformers}, and fine-tuned on the MultiNLI dataset
\citep{williams-etal-2018-broad}. Concretely, 
we choose RoBERTa, BART and DeBERTa as they cover different architectures and represent competitive approaches
across a set of NLP tasks.

\begin{table}
  \centering\small
  \setlength{\tabcolsep}{5pt}
  \begin{tabularx}{\linewidth}{lX}
    \toprule
    \textbf{Emotion} & \textbf{\emosyn} \\ 
    \cmidrule(r){1-1}\cmidrule(rl){2-2}
   anger & anger, annoyance, rage, outrage, fury, irritation \\
    \cmidrule(r){1-1}\cmidrule(rl){2-2}
   fear & fear, horror, anxiety, terror, dread, scare \\
    \cmidrule(r){1-1}\cmidrule(rl){2-2}
   joy & joy, achievement, pleasure, awesome, happy, blessed \\
    \cmidrule(r){1-1}\cmidrule(rl){2-2}
   sadness & sadness, unhappy, grief, sorrow, loneliness, depression \\
    \cmidrule(r){1-1}\cmidrule(rl){2-2}
   disgust & disgust, loathing, bitter, ugly, repugnance, revulsion \\
    \cmidrule(r){1-1}\cmidrule(rl){2-2}
   surprise & surprise, astonishment, amazement, impression, perplexity, shock \\
    \cmidrule(r){1-1}\cmidrule(rl){2-2}
   guilt & guilt, culpability, blameworthy,  responsibility, misconduct, regret \\
    \cmidrule(r){1-1}\cmidrule(rl){2-2}
   shame & shame, humiliate, embarrassment, disgrace, dishonor, discredit\\
    \bottomrule
  \end{tabularx}
  \caption{Emotion synonyms per emotion category considered in \emosyn
    prompt (details in the Appendix).}
  \label{tab:emo_syn}
\end{table} 

\paragraph{RoBERTa.} The Robustly Optimized BERT Pre-training Approach
\citep{liu2019roberta} is a modified version of BERT which includes
some changes such as the removal of the next-sentence prediction task,
the replacement of the WordPiece tokenization with a variation of the
byte-pair encoding, and the replacement of the static masking (the
same input masks are fed to the model on each epoch) with dynamic
masking (the masking is generated every time the sequence is fed to
the model). For the NLI task, we use the \textit{roberta-large-mnli}
model from Hugging Face which contains over 355M of parameters.

\begin{figure*}[t]
    \includegraphics[width=\textwidth]{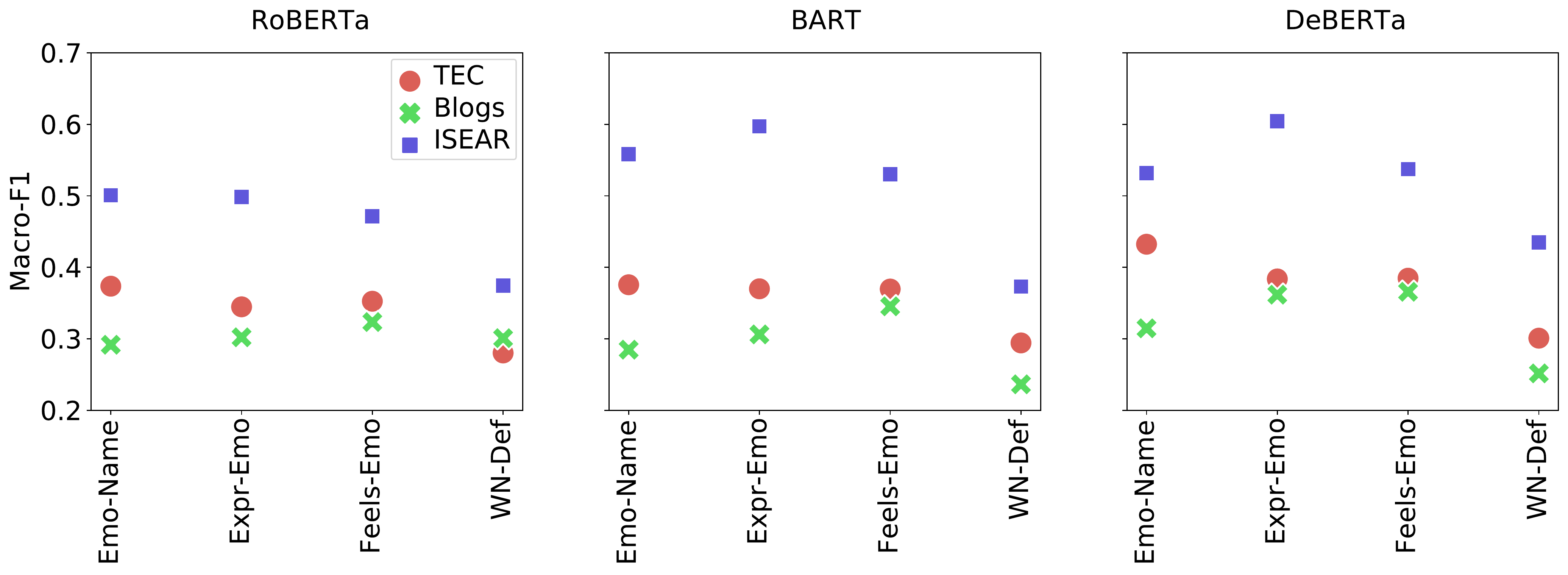}
    \caption{Results of Experiment 1. Comparison of prompts
      across NLI models and emotion datasets.}
    \label{fig:scat_main_prompts}
\end{figure*}

\paragraph{BART.} The Bidirectional and Auto-Regressive Transformer
\citep{lewis-etal-2020-bart} is a model that combines the
bidirectional encoder with an autoregressive decoder into one
sequence-to-sequence model. We use the
\textit{facebook/bart-large-mnli} model from Hugging Face with over
407M parameters.

\paragraph{DeBERTa.} The Decoding-enhanced BERT with Disentangled
Attention model \citep{he2020deberta} improves BERT and RoBERTa using
two techniques, namely disentangled attention and an enhanced mask
decoder. We use \textit{microsoft/deberta-xlarge-mnli} from Hugging
Face, which contains over 750M of parameters.

\bigskip\noindent All experiments are performed on a node equipped
with two Intel Xeon Silver 4208 CPU at 2.10GHz, 192GB RAM, as main
processors, and six GPUs NVIDIA GeForce RTX 2080Ti (with 11GB each).

\subsection{Results}

In order to answer the research questions formulated in this study, we
conduct different ZSL-NLI emotion classification experiments.

\subsubsection{Experiment 1: Are NLI models behaving the same across prompts?}

With the first experiment, we aim at observing if different NLI models
behave robustly across emotion datasets and prompts. We use each model
described in \S~\ref{sec:models} with each emotion representation
that is not a set of multiple prompts, but only consists of a single
prompt, namely \emonamme, \expremo, \feelsemo and \wordnet. We
evaluate each model using all datasets (\S~\ref{sec:datasets}).

Figure~\ref{fig:scat_main_prompts} (and Table~\ref{tab:results_models}
in the Appendix) show the results.  Each plot shows the performance of
one NLI model on the three emotion datasets using the four
prompts. We see that the performances follow the same patterns across
NLI models and emotion datasets. \emonamme is the best performing
prompt for \tec, \expremo for \isear and \feelsemo for \blogs. The
lowest performance is achieved with \wordnet.  The most successful NLI
model across the prompts is DeBERTa followed by BART and RoBERTa.

Therefore, NLI models do behave robustly across prompts. Particularly
low performance can be observed with \wordnet. This finding is in line
with previous research \citep{Yin2019}: These definitions may be
suboptimal choices, for instance, \textit{sadness} is represented via
``This person expresses emotions experienced when not in a state of
well-being''. This is ambiguous since not being in a state of
well-being may also be associated with other negative emotions such as
\textit{anger} or \textit{fear}. Interestingly, the best-performing
emotion representation on \tec is \emonamme, which resembles the
annotation procedure of just using an emotion-related hashtag for
labeling. Similarly, \expremo shows the best performance for the
self-reports of \isear (``This text expresses'') and \feelsemo on
\blogs (``This person feels''). These subtle differences in the prompt
formulations indicate that there are particular factors in the dataset
that influence the interpretation of the prompt, for instance, the
annotation procedure, the data selection or the language register
employed in the corpus, and therefore, they affect the interpretation
of the emotion by the NLI-ZSL classifier.
 
\subsubsection{Experiment 2: Should we use synonyms for emotion representation?}

In this experiment, we aim at observing whether the incorporation of
synonyms in the prompt helps the emotion interpretation. Instead of
considering only the emotion name, we use six close emotion synonyms
(see \emosyn, \exprsyn, \feelsyn in Table~\ref{tab:emo_repres} in the
Appendix).\footnote{We assume that larger numbers might show better
  performance in general, but this set of six synonyms focuses on
  close, unambiguous synonyms which undoubtedly represent the emotion in
  most contexts. We evaluate the impact of larger sets with the
  \emolex approach.} This leads to six prompts for each emotion. For
simplicity, we now only consider DeBERTa, which showed best
performances in the previous experiment.

Figure~\ref{fig:bar_synonyms} (and Table \ref{tab:results_models} in
the Appendix) shows the results of each context with just the emotion
name and with the synonyms in comparison. In general, synonym use
leads to an improvement, with some notable exceptions. For \tec, the
single use of the emotion (\emonamme) works better than using synonyms
(\emosyn). This might stem from a similarity of the prompt with the
annotation procedure, in which single hashtags were used for
labeling. Another exception is \feelsemo/\feelsyn in \blogs.
Therefore, to answer RQ2 we conclude that both context and emotion
concept representation are corpus-dependent and in some
cases synonyms support the emotion classification.

\begin{figure}
    \centering
    \includegraphics[width=0.48\textwidth]{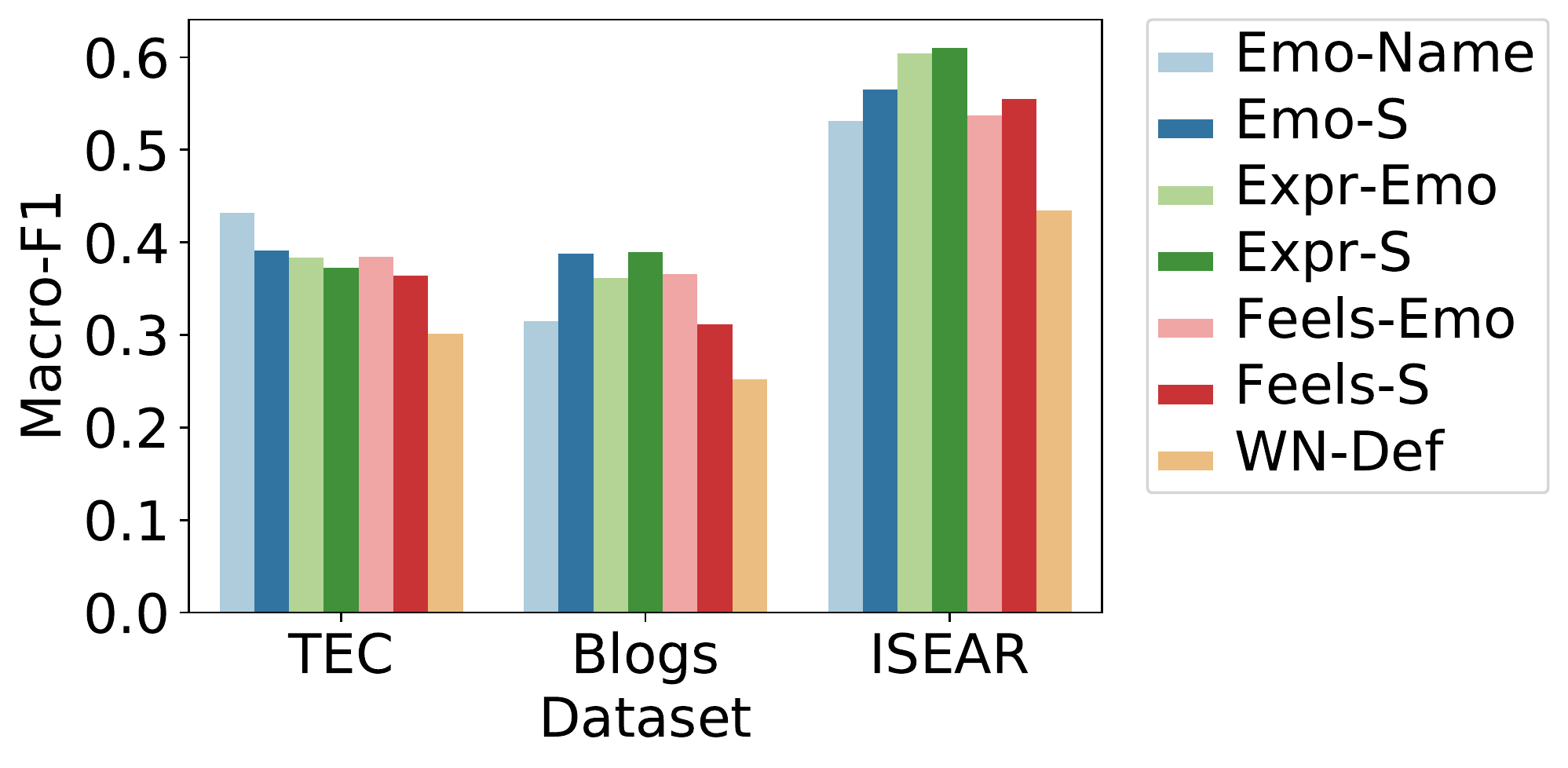}
    \caption{Results of Experiment 2. Comparison of prompts including synonym emotion
      representations across three emotion datasets (\tec, \blogs and
      \isear) using the DeBERTa model.}
    \label{fig:bar_synonyms}
\end{figure}

\subsubsection{Experiment 3: Is an ensemble of multiple prompts more
  robust across corpora?}
The previous experiments demonstrate the challenge of engineering an
emotion prompt that fits different corpora which stem from
various sources. To cope with this challenge, we analyze if the
combination of sets of prompt-generation methods in an ensemble
improves the generalizability. We use the ensemble method described in
\S~\ref{sec:ensemble_prompts} that combines the predictions given by
the set of model prompts described in \S~\ref{sec:emo_prompts} with
the DeBERTa model (d-ensemble). In addition to this realistic ensemble
model, we want to understand which performance could be achieved with
an ideal (oracle) ensemble (which we refer to as d-oracle), which
always picks the correct decision by an ensemble component, if one is
available. This serves as an upper bound and analyzes the
complementarity of the individual models.

\begin{figure}[t]
    \centering
    \includegraphics[width=0.48\textwidth]{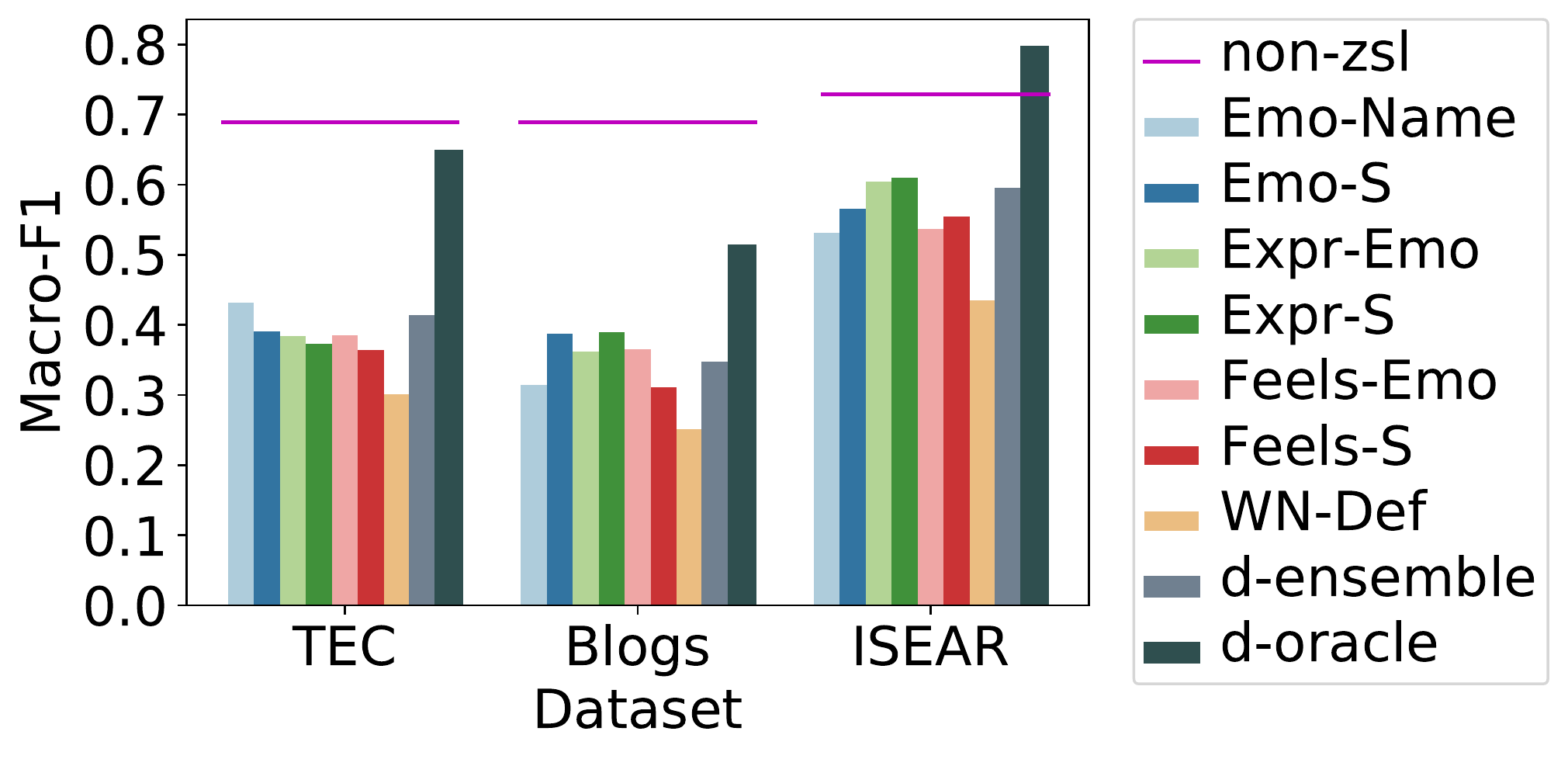}
    \caption{Results of Experiment 3. Comparison of the prompt individual models and the proposed ensemble models along with the non-zsl experiments.}
    \label{fig:bar_ensemble}
\end{figure}

Figure~\ref{fig:bar_ensemble} shows the performance for the individual
models discussed before, which participate in both the realistic and
the oracle ensemble (individual results in
Table~\ref{tab:results_models} in the Appendix, ensemble results also
in Table~\ref{tab:results_ensemble}). In addition, we see both
ensemble methods and (as a horizontal line) the supervised learning
upper bound.  We observe that the realistic ensemble (d-ensemble),
which is based on averaging the individual probabilistic outputs of
the individual models, shows a performance nearly en par with the
individual best model: For \tec, we have an \F=.41 in comparison to
the individual best \F=.43, for \blogs, we have \F=.35 in comparison
to \F=.39, and for \isear, we achieve \F=.59 in comparison to \F=.61
-- but without the necessity to pick the prompt-generating approach
beforehand or on some hold-out data.

We further see that the oracle ensemble performs better than all other
models -- this shows the variance between the models and suggests a
reason for their corpus-dependency, but also shows the potential for
other ensemble models. This oracle also approaches (or is even
slightly higher than) the supervised upper-bound. All of our current
(non-oracle) ZSL learning methods clearly underperform supervised
learning, but to various degrees. The oracle performance suggests that
sets of prompts, combined with a good ensembling method, might exist
that outperform supervised learning in emotion classification.

\begin{table}
  \centering\small
  \begin{tabularx}{\linewidth}{lX}
    \toprule
    \textbf{Emotion} & \textbf{Text} \\ 
    \cmidrule(r){1-1}\cmidrule(rl){2-2}
    anger & The sports fishermen who catch gulls instead of fish with their hooks.   It is often a mistake but it makes me angry. (\isear) \\
    \cmidrule(r){1-1}\cmidrule(rl){2-2}
    disgust & my sister got this purse, It smell like straight up KITTY LITTER. (\tec) \\
    \cmidrule(r){1-1}\cmidrule(rl){2-2}
    fear &  Oh well its nothing too too bad but its making me nervous. (\blogs)\\
    \cmidrule(r){1-1}\cmidrule(rl){2-2}
    guilt & While at primary school, I did not let a friend ring a bell although he would have liked to do it.  Afterwards I felt bad. (\isear)\\
    \cmidrule(r){1-1}\cmidrule(rl){2-2}
    joy & When I get a hug from someone I love. (\isear)\\
    \cmidrule(r){1-1}\cmidrule(rl){2-2}
    sadness & When I lost the person who meant the most to me. (\isear)\\
    \cmidrule(r){1-1}\cmidrule(rl){2-2}
    surprise &  Snow in October! (\blogs)\\
    \cmidrule(r){1-1}\cmidrule(rl){2-2}
    shame & We got into a fight with some chaps in front of our family house. The value of the property destroyed was approximately 15 000 FIM. I felt ashamed when my parents came to know about this. (\isear)\\
    \bottomrule
  \end{tabularx}
  \caption{Instances where all the prompt models agree with the
    emotion prediction.}
  \label{tab:instances}
\end{table} 

We conclude that an ensemble model is indeed more robust across
emotion datasets with different language registers and prompts, with a
performance nearly en par with the best corpus-specific prompt. This
raises the question what differences and commonalities instances have
in which models perform the same or differently. To this end, we show
examples in Table \ref{tab:instances}, in which \textit{all}
individual models did output the correct label. As we can see, these
instances contain explicit words related to the emotion conveyed. For
instance, ``lost'' for \textit{sadness}, ``love'' for \textit{joy},
``angry'' for \textit{anger}, ``nervous'' for \textit{fear},
``ashamed'' for \textit{shame}, and ``felt bad'' for
\textit{guilt}. Therefore, prompt-NLI models succeed in interpreting
emotions that are clearly expressed in the text, but vary
performance-wise when the emotion is implicitly communicated.

\subsubsection{Experiment 4: Are synonyms sufficient? Would it be even more useful to use more representations of emotions?} 

In Experiment 2 we found that the use of synonyms is beneficial in
some cases (\isear and \blogs). This raises the question if more terms
that represent the emotion would lead to an even better
performance. We evaluate this setup with the \emolex model introduced
above, in which each emotion concept is represented with a set of
prompts, where each prompt is a concept from an emotion
lexicon. Notably, in this prompt-generating methods, emotions are not
only represented by abstract emotion names or synonyms, but in
addition with (sometimes concrete) concepts, like ``gift'' or
``tears''.

\begin{table}
  \centering
  \small
  \setlength{\tabcolsep}{3pt}
  \begin{tabular}{l l ccc ccc ccc}
    \toprule
    \multirow{2}{*}{} &
      \multicolumn{3}{c}{\tec} &
      \multicolumn{3}{c}{\blogs} &
      \multicolumn{3}{c}{\isear}  \\
        \cmidrule(lr){2-4}\cmidrule(lr){5-7}\cmidrule(lr){8-10}
      Model  & P & R & F$_1$ & P & R & F$_1$ & P & R & F$_1$ \\
      \cmidrule(r){1-1}\cmidrule(lr){2-4}\cmidrule(lr){5-7}\cmidrule(l){8-10}
    d-ensemble & .42 & .44 & .41 & .40 & .65 & .35 & .67 & .62 & .59 \\ 
    d-oracle      & .63 & .69 & .65 & .51 & .80 & .51 & .82 & .80 & .80 \\
    d-emolex      & .37 & .36 & .33 & .52 & .48 & .48 & .47 & .42 & .40 \\
    \cmidrule(r){1-1}\cmidrule(lr){2-4}\cmidrule(lr){5-8}\cmidrule(l){9-10}
    non-zsl       & .69 & .69 & .69 & .72 & .71 & .69 & .73 & .73 & .73 \\
    \bottomrule
  \end{tabular}
  \caption{Results of Experiments 3 and 4. We report macro-average precision (P), macro-average recall (R), and macro-average \F (\F) for each model. d-emolex: DeBERTa using \emolex prompt, d-ensemble: ensemble model of prompts
    using DeBERTa, d-oracle: oracle ensemble model using
    DeBERTa), non-zsl: Supervised RoBERTa model fine-tune on the three
    emotion datasets.}
  \label{tab:results_ensemble}
\end{table}

Table \ref{tab:results_ensemble} shows the performance of the DeBERTa
model using the Emolex concepts (d-emolex), next to the ensemble
results. The additional concepts which cover a wide range of topics
associated with the respective emotions particularly help in the
\blogs corpus, which is the one resource that has been manually
annotated in a traditional manner. This manual annotation process
might include complex inference by the annotators to infer an emotion
category, instead of only using single words to trigger an event
description (ISEAR) or using words as hashtags (TEC). Lexicons can
therefore aid in the injection of background knowledge in the
prompt. However, this comes at the cost of considerably longer
runtimes, because the NLI models is queried for every entry in the
lexicon.

\section{Conclusion and Future Work}
We presented an analysis of various prompts for NLI-based ZSL emotion
classification. The prompts that we chose were motivated by the
various particularities of the corpora: single emotions for \tec
(tweets), ``The person feels/The text expresses'' for \blogs (blogs),
and \isear (events). In addition, we represented the emotions with
emotion names, synonyms, definitions, or with the help of
lexicons. Our experiments across these data sets showed that, to
obtain a superior performance, the prompt needs to fit well to the
corpus -- we did not find one single prompt that works well across
different corpora. To avoid the requirement for manually selecting a
prompt, we therefore devised an ensemble model that combines multiple
sets of prompts. This model is more robust and is nearly on par with
the best individual prompt. In addition, we found that representing
the emotion concept more diversely with synonyms or lexicons is
beneficial, but again corpus-specific.

Our work raises a set of future research questions. We have seen that
the oracle ensemble showed a good performance, illustrating that the
various prompts provide complementary information. This motivates
future research regarding other combination schemes, including learning a
combination based on end-to-end fine-tuned NLI
models.

We have further seen that including more concepts with the help of a
dictionary helps in one corpus, but not across corpora; however,
synonyms constantly help. This raises the question about the right
trade-off between many, but potentially inappropriate, noisy concepts
and hand-selected, high-quality concepts. A desideratum is an
automatic subselection procedure, which removes concepts that might
decrease performance and only keeps concepts that are ``compatible''
to the current language register and annotation method. Ideally, this
procedure would not make use of annotated data, because that would
limit the advantages of ZSL.

The main limitation of our current work is that we manually designed
the prompts under consideration, based on the corpora we used for
evaluation. This is a bottleneck in model development, which should
either be supported by a more guided approach which supports humans in
developing prompts, or by an automatic model that is able to
automatically generate prompts based on the language register and
concept representation in the dataset.

\section*{Acknowledgements}
We thank Enrica Troiano and Laura Oberländer for 
discussions on the topic of emotion analysis.
Roman Klinger's work is supported by the German Research Council (DFG,
project number KL 2869/1-2). Flor Miriam Plaza-del-Arco and
María-Teresa Martín Valdivia have been partially supported by the
LIVING-LANG project (RTI2018-094653-B-C21) funded by
MCIN/AEI/10.13039/501100011033 and ERDF A way of making Europe, and a
grant from the Ministry of Science, Innovation and Universities of the
Spanish Government (FPI-PRE2019-089310).

\clearpage

\onecolumn

\appendix
\label{appendix}

\section{Experiment Results}

\begin{table*}[h!]
  \centering
  \small
  \renewcommand{\arraystretch}{0.9}
  \setlength{\tabcolsep}{5pt}
  \begin{tabular}{l l ccc ccc ccc ccc}
    \toprule
    \multirow{2}{*}{} &
    \multirow{2}{*}{} &      
      \multicolumn{3}{c}{\emonamme} &
      \multicolumn{3}{c}{\expremo} &
      \multicolumn{3}{c}{\feelsemo} & 
      \multicolumn{3}{c}{\wordnet} \\
        \cmidrule(lr){3-5}\cmidrule(lr){6-8}\cmidrule(lr){9-11}\cmidrule(l){12-14}
      Dataset & Model & P & R & F$_1$ & P & R & F$_1$ & P & R & F$_1$ & P & R & F$_1$ \\
\cmidrule(l){1-1}\cmidrule(lr){2-2}\cmidrule(lr){3-5}\cmidrule(lr){6-8}\cmidrule(lr){9-11}\cmidrule(l){12-14}
                     & r & .39 & .42 & .37 
                         & .36 & .38 & .34 
                         & .39 & .40 & .35 
                         & .37 & .31 & .28 \\
            \tec      & b & .39 & .42 & .38 
                         & .38 & .42 & .37
                         & .40 & .41 & .37 
                         & .32 & .32 & .29\\
                     & d & .42 & .47 & .43 
                         & .41 & .42 & .38 
                         & .42 & .42 & .38 
                         & .44 & .33 & .30 \\
                  & d-synonyms & .42 & .42 & .39 
                                 & .39 & .40 & .37
                                 & .39 & .40 & .36
                         & --- & --- & --- \\
\cmidrule(l){1-1}\cmidrule(lr){2-2}\cmidrule(lr){3-5}\cmidrule(lr){6-8}\cmidrule(lr){9-11}\cmidrule(l){12-14}
                    & r & .32 & .62 & .29 
                        & .36 & .60 & .30 
                        & .41 & .59 & .32 
                        & .47 & .47	& .30 \\
        \blogs       & b & .33 & .58 & .28 
                        & .35 &	.62 & .31 
                        & .47 & .56 & .35 
                        & .35 & .40 & .24 \\
                   & d & .35 & .64 & .31 
                        & .41 & .65 & .36 
                        & .49 & .58 & .37 
                        & .38 & .48 & .25 \\
                   & d-synonyms & .41 & .62 & .39  
                         & .42 & .63 & .39 
                         & .36 & .60 & .31 
                         & --- & --- & ---  \\
\cmidrule(l){1-1}\cmidrule(lr){2-2}\cmidrule(lr){3-5}\cmidrule(lr){6-8}\cmidrule(lr){9-11}\cmidrule(l){12-14}          
                    & r & .58 & .50 & .50 
                        & .53 & .50 & .50 
                        & .55 & .47 & .47  
                        & .50 & .37 & .37 \\
        \isear            & b & .62 & .56 & .56
                        & .64 & .60 & .60 
                        & .68 & .53 & .53 
                        & .57 & .40 & .37 \\
                   & d & .63 & .56 & .53 
                        & .66 & .62 & .60
                        & .68 & .54	& .54
                        & .54 & .45 & .43  \\
                        & d-synonyms & .64 & .57 & .57  
                        & .64 & .62 & .61 
                        & .63 & .58 & .55 
                        & --- & --- & --- \\
    \bottomrule
  \end{tabular}
  \caption{Results from the set of prompts across
    emotion datasets (\tec, \blogs and \isear) and NLI models. We report macro-average
    precision (P), macro-average recall (R), and macro-average \F (\F)
    for each model. (r: RoBERTa, b: BART, d: DeBERTa,
    d-synonyms:
    DeBERTa using as prompts synonyms. In cases where no experiments have been performed, we use '---'. Figures \ref{fig:scat_main_prompts} and \ref{fig:bar_synonyms} in the paper depict these experiments.}
  \label{tab:results_models}
\end{table*}

\section{List of Emotion Representations as Part of Prompts}
\begin{table*}[h!]
  \centering\small
  \renewcommand{\arraystretch}{0.2}
  \begin{tabularx}{\linewidth}{lXXXp{5cm}}
    \toprule
    \textbf{Emotion} & \textbf{\emosyn} & \textbf{\exprsyn} & \textbf{\feelsyn} & \textbf{\wordnet} \\
        \cmidrule(r){1-1}\cmidrule(rl){2-2}\cmidrule(lr){3-3}\cmidrule(lr){4-4}\cmidrule(l){5-5}
    Context & $\epsilon$ & ``This text expresses\ldots'' & ``This person feels\ldots'' & ``This person expresses\ldots'' \\
    \cmidrule(r){1-1}\cmidrule(rl){2-2}\cmidrule(lr){3-3}\cmidrule(lr){4-4}\cmidrule(l){5-5}
   anger & anger, annoyance, rage, outrage, fury, irritation & anger, annoyance, rage, outrage, fury, irritation & anger, annoyed, rage, outraged, furious, irritated & a strong feeling of annoyance, displeasure, or hostility \\
    \cmidrule(r){1-1}\cmidrule(rl){2-2}\cmidrule(lr){3-3}\cmidrule(lr){4-4}\cmidrule(l){5-5}
   fear & fear, horror, anxiety, terror, dread, scare & fear, horror, anxiety, terror, dread, scare & fear, horror, anxiety, terrified, dread, scared & an unpleasant emotion caused by the belief that someone or something is dangerous, likely to cause pain , or a threat\\
    \cmidrule(r){1-1}\cmidrule(rl){2-2}\cmidrule(lr){3-3}\cmidrule(lr){4-4}\cmidrule(l){5-5}
   joy & joy, achievement, pleasure, awesome, happy, blessed & joy, an achievement, pleasure, the awesome, happiness, the blessing & joyful, accomplished,  pleasure, awesome, happy, blessed & a feeling of great pleasure and happiness\\
    \cmidrule(r){1-1}\cmidrule(rl){2-2}\cmidrule(lr){3-3}\cmidrule(lr){4-4}\cmidrule(l){5-5}
   sadness & sadness, unhappy, grief, sorrow, loneliness, depression & sadness, unhappiness, grief, sorrow, loneliness, depression & sadness, unhappy, grieved, sorrow, lonely, depression & emotions experienced when not in a state of well-being\\
    \cmidrule(r){1-1}\cmidrule(rl){2-2}\cmidrule(lr){3-3}\cmidrule(lr){4-4}\cmidrule(l){5-5}
   disgust & disgust, loathing, bitter, ugly, repugnance, revulsion & disgust, loathing, bitterness, ugliness, repugnance, revulsion & disgusted, loathing, bitter, ugly, repugnance, revulsion & a feeling of revulsion or strong disapproval aroused by something unpleasant or offensive\\
    \cmidrule(r){1-1}\cmidrule(rl){2-2}\cmidrule(lr){3-3}\cmidrule(lr){4-4}\cmidrule(l){5-5}
   surprise & surprise, astonishment, amazement, impression, perplexity, shock & surprise, astonishment, amazement, impression, perplexity, shock & surprised, astonishment, amazement, impressed, perplexed, shocked & a feeling of mild astonishment or shock caused by something unexpected \\
    \cmidrule(r){1-1}\cmidrule(rl){2-2}\cmidrule(lr){3-3}\cmidrule(lr){4-4}\cmidrule(l){5-5}
   guilt & guilt, culpability, blameworthy,  responsibility, misconduct, regret & guilt, culpability, responsibility, blameworthy, misconduct, regret & guilty, culpable, responsible, blame, misconduct, regretful & a feeling of having done wrong or failed in an obligation\\
    \cmidrule(r){1-1}\cmidrule(rl){2-2}\cmidrule(lr){3-3}\cmidrule(lr){4-4}\cmidrule(l){5-5}
   shame & shame, humiliate, embarrassment, disgrace, dishonor, discredit & shame, humiliation, embarrassment, disgrace, dishonor, discredit &  shameful, humiliated, embarrassed, disgraced, dishonored, discredit & a painful feeling of humiliation or distress caused by the consciousness of wrong or foolish behavior \\
    \bottomrule
  \end{tabularx}
  \caption{Emotion representation in prompts \emosyn, \exprsyn,
    \feelsyn, and \wordnet.}
  \label{tab:emo_repres}
\end{table*} 


\begin{thebibliography}{49}
\expandafter\ifx\csname natexlab\endcsname\relax\def\natexlab#1{#1}\fi

\bibitem[{Aman and Szpakowicz(2007)}]{aman2007identifying}
Saima Aman and Stan Szpakowicz. 2007.
\newblock \href {https://doi.org/10.1007/978-3-540-74628-7\_27} {{Identifying
  Expressions of Emotion in Text}}.
\newblock In \emph{Text, Speech and Dialogue, 10th International Conference,
  {TSD} 2007, Pilsen, Czech Republic, September 3-7, 2007, Proceedings}, volume
  4629 of \emph{Lecture Notes in Computer Science}, pages 196--205. Springer.

\bibitem[{Basile et~al.(2021)Basile, P{\'e}rez-Torr{\'o}, and
  Franco-Salvador}]{basile-etal-2021-probabilistic}
Angelo Basile, Guillermo P{\'e}rez-Torr{\'o}, and Marc Franco-Salvador. 2021.
\newblock \href {https://aclanthology.org/2021.ranlp-main.16.pdf}
  {{"Probabilistic Ensembles of Zero- and Few-Shot Learning Models for Emotion
  Classification"}}.
\newblock In \emph{Proceedings of the International Conference on Recent
  Advances in Natural Language Processing (RANLP 2021)}, pages 128--137, Held
  Online. INCOMA Ltd.

\bibitem[{Bostan et~al.(2020)Bostan, Kim, and
  Klinger}]{bostan-etal-2020-goodnewseveryone}
Laura Ana~Maria Bostan, Evgeny Kim, and Roman Klinger. 2020.
\newblock \href {https://aclanthology.org/2020.lrec-1.194}
  {{"{G}ood{N}ews{E}veryone: A Corpus of News Headlines Annotated with
  Emotions, Semantic Roles, and Reader Perception"}}.
\newblock In \emph{Proceedings of the 12th Language Resources and Evaluation
  Conference}, pages 1554--1566, Marseille, France. European Language Resources
  Association.

\bibitem[{Bostan and Klinger(2018)}]{bostan-klinger-2018-analysis}
Laura-Ana-Maria Bostan and Roman Klinger. 2018.
\newblock \href {https://aclanthology.org/C18-1179} {{"An Analysis of Annotated
  Corpora for Emotion Classification in Text"}}.
\newblock In \emph{Proceedings of the 27th International Conference on
  Computational Linguistics}, pages 2104--2119, Santa Fe, New Mexico, USA.
  Association for Computational Linguistics.

\bibitem[{Brown et~al.(2020)Brown, Mann, Ryder, Subbiah, Kaplan, Dhariwal,
  Neelakantan, Shyam, Sastry, Askell, Agarwal, Herbert{-}Voss, Krueger,
  Henighan, Child, Ramesh, Ziegler, Wu, Winter, Hesse, Chen, Sigler, Litwin,
  Gray, Chess, Clark, Berner, McCandlish, Radford, Sutskever, and
  Amodei}]{BrownMRSKDNSSAA20}
Tom~B. Brown, Benjamin Mann, Nick Ryder, Melanie Subbiah, Jared Kaplan,
  Prafulla Dhariwal, Arvind Neelakantan, Pranav Shyam, Girish Sastry, Amanda
  Askell, Sandhini Agarwal, Ariel Herbert{-}Voss, Gretchen Krueger, Tom
  Henighan, Rewon Child, Aditya Ramesh, Daniel~M. Ziegler, Jeffrey Wu, Clemens
  Winter, Christopher Hesse, Mark Chen, Eric Sigler, Mateusz Litwin, Scott
  Gray, Benjamin Chess, Jack Clark, Christopher Berner, Sam McCandlish, Alec
  Radford, Ilya Sutskever, and Dario Amodei. 2020.
\newblock \href
  {https://proceedings.neurips.cc/paper/2020/hash/1457c0d6bfcb4967418bfb8ac142f64a-Abstract.html}
  {{Language Models are Few-Shot Learners}}.
\newblock In \emph{Advances in Neural Information Processing Systems 33: Annual
  Conference on Neural Information Processing Systems 2020, NeurIPS 2020,
  December 6-12, 2020, virtual}.

\bibitem[{Buechel and Hahn(2017)}]{buechel-hahn-2017-emobank}
Sven Buechel and Udo Hahn. 2017.
\newblock \href {https://aclanthology.org/E17-2092} {{"{E}mo{B}ank: Studying
  the Impact of Annotation Perspective and Representation Format on Dimensional
  Emotion Analysis"}}.
\newblock In \emph{Proceedings of the 15th Conference of the {E}uropean Chapter
  of the Association for Computational Linguistics: Volume 2, Short Papers},
  pages 578--585, Valencia, Spain. Association for Computational Linguistics.

\bibitem[{Buechel et~al.(2021)Buechel, Modersohn, and Hahn}]{Buechel2021}
Sven Buechel, Luise Modersohn, and Udo Hahn. 2021.
\newblock \href {https://aclanthology.org/2021.emnlp-main.728} {{"Towards
  Label-Agnostic Emotion Embeddings"}}.
\newblock In \emph{Proceedings of the 2021 Conference on Empirical Methods in
  Natural Language Processing}, pages 9231--9249, Online and Punta Cana,
  Dominican Republic. Association for Computational Linguistics.

\bibitem[{Devlin et~al.(2019)Devlin, Chang, Lee, and Toutanova}]{Devlin2019}
Jacob Devlin, Ming-Wei Chang, Kenton Lee, and Kristina Toutanova. 2019.
\newblock \href {https://aclanthology.org/N19-1423.pdf} {{"{BERT}: Pre-training
  of Deep Bidirectional Transformers for Language Understanding"}}.
\newblock In \emph{Proceedings of the 2019 Conference of the North {A}merican
  Chapter of the Association for Computational Linguistics: Human Language
  Technologies, Volume 1 (Long and Short Papers)}, pages 4171--4186,
  Minneapolis, Minnesota. Association for Computational Linguistics.

\bibitem[{D'mello and Graesser(2007)}]{d2007mind}
Sidney D'mello and Arthur Graesser. 2007.
\newblock \href {https://dl.acm.org/doi/10.5555/1563601.1563631} {Mind and
  body: {Dialogue} and posture for affect detection in learning environments}.
\newblock In \emph{Proceedings of the 2007 conference on Artificial
  Intelligence in Education: Building Technology Rich Learning Contexts That
  Work}, page 161–168, NLD. IOS Press.

\bibitem[{Doan~Dang et~al.(2021)Doan~Dang, Oberl{\"{a}}nder, and
  Klinger}]{DoanDang2021}
Bao~Minh Doan~Dang, Laura Oberl{\"{a}}nder, and Roman Klinger. 2021.
\newblock \href
  {https://konvens2021.phil.hhu.de/wp-content/uploads/2021/09/2021.KONVENS-1.7.pdf}
  {{Emotion Stimulus Detection in German News Headlines}}.
\newblock In \emph{Proceedings of the 17th Conference on Natural Language
  Processing (KONVENS 2021)}, D\"usseldorf, Germany. German Society for
  Computational Linguistics \& Language Technology.

\bibitem[{Ekman(1992)}]{ekman1992argument}
Paul Ekman. 1992.
\newblock \href {https://doi.org/10.1080/02699939208411068} {An argument for
  basic emotions}.
\newblock \emph{Cognition and Emotion}, 6(3-4):169--200.

\bibitem[{Haider et~al.(2020)Haider, Eger, Kim, Klinger, and
  Menninghaus}]{haider-etal-2020-po}
Thomas Haider, Steffen Eger, Evgeny Kim, Roman Klinger, and Winfried
  Menninghaus. 2020.
\newblock \href {https://aclanthology.org/2020.lrec-1.205} {{"{PO}-{EMO}:
  Conceptualization, Annotation, and Modeling of Aesthetic Emotions in {G}erman
  and {E}nglish Poetry"}}.
\newblock In \emph{Proceedings of the 12th Language Resources and Evaluation
  Conference}, pages 1652--1663, Marseille, France. European Language Resources
  Association.

\bibitem[{Halder et~al.(2020)Halder, Akbik, Krapac, and
  Vollgraf}]{halder-etal-2020-task}
Kishaloy Halder, Alan Akbik, Josip Krapac, and Roland Vollgraf. 2020.
\newblock \href {https://aclanthology.org/2020.coling-main.285} {{"Task-Aware
  Representation of Sentences for Generic Text Classification"}}.
\newblock In \emph{Proceedings of the 28th International Conference on
  Computational Linguistics}, pages 3202--3213, Barcelona, Spain (Online).
  International Committee on Computational Linguistics.

\bibitem[{He et~al.(2021)He, Liu, Gao, and Chen}]{he2020deberta}
Pengcheng He, Xiaodong Liu, Jianfeng Gao, and Weizhu Chen. 2021.
\newblock \href {https://openreview.net/forum?id=XPZIaotutsD} {{DeBERTa:
  Decoding-enhanced BERT with Disentangled Attention}}.
\newblock In \emph{International Conference on Learning Representations}.

\bibitem[{Hofmann et~al.(2020)Hofmann, Troiano, Sassenberg, and
  Klinger}]{Hofmann2020}
Jan Hofmann, Enrica Troiano, Kai Sassenberg, and Roman Klinger. 2020.
\newblock \href {https://aclanthology.org/2020.coling-main.11} {{"Appraisal
  Theories for Emotion Classification in Text"}}.
\newblock In \emph{Proceedings of the 28th International Conference on
  Computational Linguistics}, Barcelona, Spain (Online). International
  Committee on Computational Linguistics.

\bibitem[{K{\"o}per et~al.(2017)K{\"o}per, Kim, and
  Klinger}]{koper-etal-2017-ims}
Maximilian K{\"o}per, Evgeny Kim, and Roman Klinger. 2017.
\newblock \href {https://aclanthology.org/W17-5206} {{"{IMS} at
  {E}mo{I}nt-2017: Emotion Intensity Prediction with Affective Norms,
  Automatically Extended Resources and Deep Learning"}}.
\newblock In \emph{Proceedings of the 8th Workshop on Computational Approaches
  to Subjectivity, Sentiment and Social Media Analysis}, pages 50--57,
  Copenhagen, Denmark. Association for Computational Linguistics.

\bibitem[{Lazaridou et~al.(2015)Lazaridou, Dinu, and Baroni}]{Lazaridou2015}
Angeliki Lazaridou, Georgiana Dinu, and Marco Baroni. 2015.
\newblock \href {https://aclanthology.org/P15-1027} {{"Hubness and Pollution:
  Delving into Cross-Space Mapping for Zero-Shot Learning"}}.
\newblock In \emph{Proceedings of the 53rd Annual Meeting of the Association
  for Computational Linguistics and the 7th International Joint Conference on
  Natural Language Processing (Volume 1: Long Papers)}, pages 270--280,
  Beijing, China. Association for Computational Linguistics.

\bibitem[{Lewis et~al.(2020)Lewis, Liu, Goyal, Ghazvininejad, Mohamed, Levy,
  Stoyanov, and Zettlemoyer}]{lewis-etal-2020-bart}
Mike Lewis, Yinhan Liu, Naman Goyal, Marjan Ghazvininejad, Abdelrahman Mohamed,
  Omer Levy, Veselin Stoyanov, and Luke Zettlemoyer. 2020.
\newblock \href {https://aclanthology.org/2020.acl-main.703} {{{BART}:
  Denoising Sequence-to-Sequence Pre-training for Natural Language Generation,
  Translation, and Comprehension}}.
\newblock In \emph{Proceedings of the 58th Annual Meeting of the Association
  for Computational Linguistics}, pages 7871--7880, Online. Association for
  Computational Linguistics.

\bibitem[{Li et~al.(2021)Li, Gao, Feng, Zhang, and
  Wang}]{li-etal-2021-boundary}
Xiangju Li, Wei Gao, Shi Feng, Yifei Zhang, and Daling Wang. 2021.
\newblock \href {https://doi.org/10.18653/v1/2021.findings-acl.60} {{"Boundary
  Detection with {BERT} for Span-level Emotion Cause Analysis"}}.
\newblock In \emph{Findings of the Association for Computational Linguistics:
  ACL-IJCNLP 2021}, pages 676--682, Online. Association for Computational
  Linguistics.

\bibitem[{Liu et~al.(2020)Liu, Ott, Goyal, Du, Joshi, Chen, Levy, Lewis,
  Zettlemoyer, and Stoyanov}]{liu2019roberta}
Yinhan Liu, Myle Ott, Naman Goyal, Jingfei Du, Mandar Joshi, Danqi Chen, Omer
  Levy, Mike Lewis, Luke Zettlemoyer, and Veselin Stoyanov. 2020.
\newblock \href {https://openreview.net/forum?id=SyxS0T4tvS} {{Ro{BERT}a: A
  Robustly Optimized {BERT} Pretraining Approach}}.

\bibitem[{Menninghaus et~al.(2019)Menninghaus, Wagner, Wassiliwizky, Schindler,
  Hanich, Jacobsen, and Koelsch}]{Menninghaus2019}
Winfried Menninghaus, Valentin Wagner, Eugen Wassiliwizky, Ines Schindler,
  Julian Hanich, Thomas Jacobsen, and Stefan Koelsch. 2019.
\newblock \href {https://pubmed.ncbi.nlm.nih.gov/30802122/} {{{W}hat are
  aesthetic emotions?}}
\newblock \emph{Psychological Review}, 126(2):171--195.

\bibitem[{Miller(1995)}]{10.1145/219717.219748}
George~A. Miller. 1995.
\newblock \href {https://doi.org/10.1145/219717.219748} {Wordnet: A lexical
  database for english}.
\newblock \emph{Commun. ACM}, 38(11):39–41.

\bibitem[{Mohammad(2012)}]{mohammad2012emotional}
Saif Mohammad. 2012.
\newblock \href {https://aclanthology.org/volumes/S12-1} {{\# Emotional
  tweets}}.
\newblock In \emph{* SEM 2012: The First Joint Conference on Lexical and
  Computational Semantics--Volume 1: Proceedings of the main conference and the
  shared task, and Volume 2: Proceedings of the Sixth International Workshop on
  Semantic Evaluation (SemEval 2012)}, pages 246--255.

\bibitem[{Mohammad and
  Bravo-Marquez(2017)}]{mohammad-bravo-marquez-2017-emotion}
Saif Mohammad and Felipe Bravo-Marquez. 2017.
\newblock \href {https://aclanthology.org/S17-1007} {{Emotion Intensities in
  Tweets}}.
\newblock In \emph{Proceedings of the 6th Joint Conference on Lexical and
  Computational Semantics (*{SEM} 2017)}, pages 65--77, Vancouver, Canada.
  Association for Computational Linguistics.

\bibitem[{Mohammad et~al.(2018{\natexlab{a}})Mohammad, Bravo-Marquez, Salameh,
  and Kiritchenko}]{mohammad-etal-2018-semeval}
Saif Mohammad, Felipe Bravo-Marquez, Mohammad Salameh, and Svetlana
  Kiritchenko. 2018{\natexlab{a}}.
\newblock \href {https://aclanthology.org/S18-1001} {{S}em{E}val-2018 task 1:
  Affect in tweets}.
\newblock In \emph{Proceedings of The 12th International Workshop on Semantic
  Evaluation}, pages 1--17, New Orleans, Louisiana. Association for
  Computational Linguistics.

\bibitem[{Mohammad et~al.(2018{\natexlab{b}})Mohammad, Bravo-Marquez, Salameh,
  and Kiritchenko}]{Mohammad2018}
Saif Mohammad, Felipe Bravo-Marquez, Mohammad Salameh, and Svetlana
  Kiritchenko. 2018{\natexlab{b}}.
\newblock \href {https://doi.org/10.18653/v1/S18-1001} {{S}em{E}val-2018 task
  1: Affect in tweets}.
\newblock In \emph{Proceedings of The 12th International Workshop on Semantic
  Evaluation}, pages 1--17, New Orleans, Louisiana. Association for
  Computational Linguistics.

\bibitem[{Mohammad and Turney(2013)}]{mohammad2013crowdsourcing}
Saif~M Mohammad and Peter~D Turney. 2013.
\newblock \href
  {http://www.saifmohammad.com/WebDocs/Crowdsourcing-MohammadTurney-CI.pdf}
  {Crowdsourcing a word--emotion association lexicon}.
\newblock \emph{Computational intelligence}, 29(3):436--465.

\bibitem[{Park et~al.(2021)Park, Kim, Ye, Jeon, Park, and
  Oh}]{park-etal-2021-dimensional}
Sungjoon Park, Jiseon Kim, Seonghyeon Ye, Jaeyeol Jeon, Hee~Young Park, and
  Alice Oh. 2021.
\newblock \href {https://aclanthology.org/2021.emnlp-main.358} {{Dimensional
  Emotion Detection from Categorical Emotion}}.
\newblock In \emph{Proceedings of the 2021 Conference on Empirical Methods in
  Natural Language Processing}, pages 4367--4380, Online and Punta Cana,
  Dominican Republic. Association for Computational Linguistics.

\bibitem[{Plaza-del Arco et~al.(2021)Plaza-del Arco, Jiménez~Zafra,
  Montejo~Ráez, Molina~González, Ureña~López, and
  Martín~Valdivia}]{plaza2021emoevales}
Flor~Miriam Plaza-del Arco, Salud~M. Jiménez~Zafra, Arturo Montejo~Ráez,
  M.~Dolores Molina~González, Luis~Alfonso Ureña~López, and María~Teresa
  Martín~Valdivia. 2021.
\newblock \href
  {http://journal.sepln.org/sepln/ojs/ojs/index.php/pln/article/view/6385}
  {{Overview of the EmoEvalEs task on emotion detection for Spanish at IberLEF
  2021}}.
\newblock \emph{Procesamiento del Lenguaje Natural}, 67(3):273--294.

\bibitem[{{Plaza-del-Arco} et~al.(2020){Plaza-del-Arco}, Strapparava,
  {Ureña-López}, and {Martín-Valdivia}}]{plaza-del-arco-etal-2020-emoevent}
{Flor Miriam} {Plaza-del-Arco}, Carlo Strapparava, L.~Alfonso {Ureña-López},
  and M.~Teresa {Martín-Valdivia}. 2020.
\newblock \href {https://aclanthology.org/2020.lrec-1.186} {{E}mo{E}vent: A
  {M}ultilingual {E}motion {C}orpus based on different {E}vents}.
\newblock In \emph{Proceedings of the 12th Language Resources and Evaluation
  Conference}, pages 1492--1498, Marseille, France. European Language Resources
  Association.

\bibitem[{Plutchik(2001)}]{plutchik2001nature}
Robert Plutchik. 2001.
\newblock \href {https://www.jstor.org/stable/27857503} {{The Nature of
  Emotions: Human emotions have deep evolutionary roots, a fact that may
  explain their complexity and provide tools for clinical practice}}.
\newblock \emph{American scientist}, 89(4):344--350.

\bibitem[{Preo{\c{t}}iuc-Pietro et~al.(2016)Preo{\c{t}}iuc-Pietro, Schwartz,
  Park, Eichstaedt, Kern, Ungar, and
  Shulman}]{preotiuc-pietro-etal-2016-modelling}
Daniel Preo{\c{t}}iuc-Pietro, H.~Andrew Schwartz, Gregory Park, Johannes
  Eichstaedt, Margaret Kern, Lyle Ungar, and Elisabeth Shulman. 2016.
\newblock \href {https://aclanthology.org/W16-0404} {{"Modelling Valence and
  Arousal in {F}acebook posts"}}.
\newblock In \emph{Proceedings of the 7th Workshop on Computational Approaches
  to Subjectivity, Sentiment and Social Media Analysis}, pages 9--15, San
  Diego, California. Association for Computational Linguistics.

\bibitem[{Radovanovic; et~al.(2010)Radovanovic;, Nanopoulos, and
  Ivanovic}]{radovanovic10a}
Milos Radovanovic;, Alexandros Nanopoulos, and Mirjana Ivanovic. 2010.
\newblock \href {http://jmlr.org/papers/v11/radovanovic10a.html} {{Hubs in
  Space: Popular Nearest Neighbors in High-Dimensional Data}}.
\newblock \emph{Journal of Machine Learning Research}, 11(86):2487--2531.

\bibitem[{Rios and Kavuluru(2018)}]{rios-kavuluru-2018-shot}
Anthony Rios and Ramakanth Kavuluru. 2018.
\newblock \href {https://aclanthology.org/D18-1352} {{"Few-Shot and Zero-Shot
  Multi-Label Learning for Structured Label Spaces"}}.
\newblock In \emph{Proceedings of the 2018 Conference on Empirical Methods in
  Natural Language Processing}, pages 3132--3142, Brussels, Belgium.
  Association for Computational Linguistics.

\bibitem[{Russell and Mehrabian(1977)}]{russell1977evidence}
James~A Russell and Albert Mehrabian. 1977.
\newblock \href {https://doi.org/https://doi.org/10.1016/0092-6566(77)90037-X}
  {Evidence for a three-factor theory of emotions}.
\newblock \emph{Journal of research in Personality}, 11(3):273--294.

\bibitem[{Sappadla et~al.(2016)Sappadla, Nam, Loza~Menc{\'{\i}}a, and
  F{\"{u}}rnkranz}]{esann16zeroshot}
Prateek~Veeranna Sappadla, Jinseok Nam, Eneldo Loza~Menc{\'{\i}}a, and Johannes
  F{\"{u}}rnkranz. 2016.
\newblock \href
  {https://www.esann.org/sites/default/files/proceedings/legacy/es2016-174.pdf}
  {{Using Semantic Similarity for Multi-Label Zero-Shot Classification of Text
  Documents}}.
\newblock In \emph{Proceedings of the 23rd European Symposium on Artificial
  Neural Networks, Computational Intelligence and Machine Learning (ESANN-16)},
  Bruges, Belgium. d-side publications.

\bibitem[{Scherer and Wallbott(1997)}]{Scherer1997}
Klaus~R. Scherer and Harald~G. Wallbott. 1997.
\newblock \href
  {https://www.unige.ch/cisa/research/materials-and-online-research/research-material/}
  {{The {ISEAR} Questionnaire and Codebook}}.
\newblock Geneva Emotion Research Group.

\bibitem[{Smith and Ellsworth(1985)}]{smith1985patterns}
Craig~A. Smith and Phoebe~C. Ellsworth. 1985.
\newblock \href {https://psycnet.apa.org/record/1985-19287-001} {Patterns of
  cognitive appraisal in emotion.}
\newblock \emph{Journal of personality and social psychology}, 48(4):813.

\bibitem[{Socher et~al.(2013)Socher, Ganjoo, Manning, and Ng}]{2999611.2999716}
Richard Socher, Milind Ganjoo, Christopher~D. Manning, and Andrew~Y. Ng. 2013.
\newblock \href {https://dl.acm.org/doi/10.5555/2999611.2999716} {{Zero-Shot
  Learning through Cross-Modal Transfer}}.
\newblock In \emph{Proceedings of the 26th International Conference on Neural
  Information Processing Systems - Volume 1}, NIPS'13, page 935–943, Red
  Hook, NY, USA. Curran Associates Inc.

\bibitem[{Sreeja and Mahalaksmi(2015)}]{sreeja2015applying}
P.S. Sreeja and G.S. Mahalaksmi. 2015.
\newblock \href
  {https://www.thefreelibrary.com/Applying+vector+space+model+for+poetic+emotion+recognition.-a0420324663}
  {Applying vector space model for poetic emotion recognition}.
\newblock \emph{Advances in Natural and Applied Sciences}, 9(6 SE):486--491.

\bibitem[{Stranisci et~al.(2022)Stranisci, Frenda, Ceccaldi, Basile, Damiano,
  and Patti}]{stranisci-EtAl:2022:LREC}
Marco~Antonio Stranisci, Simona Frenda, Eleonora Ceccaldi, Valerio Basile,
  Rossana Damiano, and Viviana Patti. 2022.
\newblock \href {https://aclanthology.org/2022.lrec-1.406} {Appreddit: a corpus
  of reddit posts annotated for appraisal}.
\newblock In \emph{Proceedings of the Language Resources and Evaluation
  Conference}, pages 3809--3818, Marseille, France. European Language Resources
  Association.

\bibitem[{Tafreshi et~al.(2021)Tafreshi, De~Clercq, Barriere, Buechel, Sedoc,
  and Balahur}]{tafreshi-etal-2021-wassa}
Shabnam Tafreshi, Orphee De~Clercq, Valentin Barriere, Sven Buechel, Jo{\~a}o
  Sedoc, and Alexandra Balahur. 2021.
\newblock \href {https://aclanthology.org/2021.wassa-1.10} {{{"WASSA} 2021
  Shared Task: Predicting Empathy and Emotion in Reaction to News Stories"}}.
\newblock In \emph{Proceedings of the Eleventh Workshop on Computational
  Approaches to Subjectivity, Sentiment and Social Media Analysis}, pages
  92--104, Online. Association for Computational Linguistics.

\bibitem[{Troiano et~al.(2023)Troiano, Oberl\"ander, and Klinger}]{Troiano2023}
Enrica Troiano, Laura Oberl\"ander, and Roman Klinger. 2023.
\newblock \href {https://doi.org/10.1162/coli_a_00461} {Dimensional modeling of
  emotions in text with appraisal theories: Corpus creation, annotation
  reliability, and prediction}.
\newblock \emph{Computational Linguistics}, 49(1).

\bibitem[{Troiano et~al.(2019)Troiano, Pad{\'o}, and Klinger}]{Troiano2019}
Enrica Troiano, Sebastian Pad{\'o}, and Roman Klinger. 2019.
\newblock \href {https://aclanthology.org/P19-1391} {{"Crowdsourcing and
  Validating Event-focused Emotion Corpora for {G}erman and {E}nglish"}}.
\newblock In \emph{Proceedings of the 57th Annual Meeting of the Association
  for Computational Linguistics}, Florence, Italy. Association for
  Computational Linguistics.

\bibitem[{Wang et~al.(2019)Wang, Zheng, Yu, and Miao}]{10.1145/3293318}
Wei Wang, Vincent~W. Zheng, Han Yu, and Chunyan Miao. 2019.
\newblock \href {https://doi.org/10.1145/3293318} {{A Survey of Zero-Shot
  Learning: Settings, Methods, and Applications}}.
\newblock \emph{ACM Trans. Intell. Syst. Technol.}, 10(2).

\bibitem[{Williams et~al.(2018)Williams, Nangia, and
  Bowman}]{williams-etal-2018-broad}
Adina Williams, Nikita Nangia, and Samuel Bowman. 2018.
\newblock \href {https://aclanthology.org/N18-1101} {{"A Broad-Coverage
  Challenge Corpus for Sentence Understanding through Inference"}}.
\newblock In \emph{Proceedings of the 2018 Conference of the North {A}merican
  Chapter of the Association for Computational Linguistics: Human Language
  Technologies, Volume 1 (Long Papers)}, pages 1112--1122, New Orleans,
  Louisiana. Association for Computational Linguistics.

\bibitem[{Wolf et~al.(2020)Wolf, Debut, Sanh, Chaumond, Delangue, Moi, Cistac,
  Rault, Louf, Funtowicz, Davison, Shleifer, von Platen, Ma, Jernite, Plu, Xu,
  Le~Scao, Gugger, Drame, Lhoest, and Rush}]{wolf-etal-2020-transformers}
Thomas Wolf, Lysandre Debut, Victor Sanh, Julien Chaumond, Clement Delangue,
  Anthony Moi, Pierric Cistac, Tim Rault, Remi Louf, Morgan Funtowicz, Joe
  Davison, Sam Shleifer, Patrick von Platen, Clara Ma, Yacine Jernite, Julien
  Plu, Canwen Xu, Teven Le~Scao, Sylvain Gugger, Mariama Drame, Quentin Lhoest,
  and Alexander Rush. 2020.
\newblock \href {https://aclanthology.org/2020.emnlp-demos.6} {{"Transformers:
  State-of-the-Art Natural Language Processing"}}.
\newblock In \emph{Proceedings of the 2020 Conference on Empirical Methods in
  Natural Language Processing: System Demonstrations}, Online. Association for
  Computational Linguistics.

\bibitem[{Xian et~al.(2019)Xian, Lampert, Schiele, and Akata}]{8413121}
Yongqin Xian, Christoph~H. Lampert, Bernt Schiele, and Zeynep Akata. 2019.
\newblock \href {https://doi.org/10.1109/TPAMI.2018.2857768} {{Zero-Shot
  Learning—A Comprehensive Evaluation of the Good, the Bad and the Ugly}}.
\newblock \emph{IEEE Transactions on Pattern Analysis and Machine
  Intelligence}, 41(9):2251--2265.

\bibitem[{Yin et~al.(2019)Yin, Hay, and Roth}]{Yin2019}
Wenpeng Yin, Jamaal Hay, and Dan Roth. 2019.
\newblock \href {https://aclanthology.org/D19-1404.pdf} {{"Benchmarking
  Zero-shot Text Classification: Datasets, Evaluation and Entailment
  Approach"}}.
\newblock In \emph{Proceedings of the 2019 Conference on Empirical Methods in
  Natural Language Processing and the 9th International Joint Conference on
  Natural Language Processing (EMNLP-IJCNLP)}, pages 3914--3923, Hong Kong,
  China. Association for Computational Linguistics.

\end{thebibliography}
\end{document}